\pdfoutput=1

\documentclass[11pt]{article}

\usepackage[]{acl}

\usepackage{times}
\usepackage{latexsym}

\usepackage[T1]{fontenc}

\usepackage[utf8]{inputenc}

\usepackage{microtype}

\usepackage{amsmath}
\usepackage{algorithmicx}
\usepackage{algorithm}
\usepackage{amssymb}
\usepackage{graphicx}
\usepackage{subfigure}
\usepackage{multirow}
\usepackage{booktabs}
\usepackage{bbm}
\usepackage{bm}
\usepackage{algpseudocode}

\title{Detect Rumors in Microblog Posts for Low-Resource Domains via Adversarial Contrastive Learning}

\author{Hongzhan Lin$^{1,2}$, Jing Ma$^{2,*}$, Liangliang Chen$^1$, Zhiwei Yang$^3$, Mingfei Cheng$^1$, Guang Chen$^{1,}$\thanks{{\; Corresponding authors.}} \\
        $^1$Beijing University of Posts and Telecommunications, Beijing, China \\ $^2$Hong Kong Baptist University, Hong Kong SAR, China \\ $^3$Jilin University, Changchun, China \\
        \texttt{\{linhongzhan,outside,mingfeicheng,chenguang\}@bupt.edu.cn}\\
        \texttt{majing@comp.hkbu.edu.hk, yangzw18@mails.jlu.edu.cn}}

\begin{document}
\maketitle
\begin{abstract}
Massive false rumors emerging along with breaking news or trending topics severely hinder the truth. Existing rumor detection approaches achieve promising performance on the yesterday's news, since there is enough corpus collected from the same domain for model training. However, they are poor at detecting rumors about unforeseen events especially those propagated in different languages due to the lack of training data and prior knowledge (i.e., low-resource regimes). In this paper, we propose an adversarial contrastive learning framework to detect rumors by adapting the features learned from well-resourced rumor data to that of the low-resourced. Our model explicitly overcomes the restriction of domain and/or language usage via language alignment and a novel supervised contrastive training paradigm. Moreover, we develop an adversarial augmentation mechanism to further enhance the robustness of low-resource rumor representation. Extensive experiments conducted on two low-resource datasets collected from real-world microblog platforms demonstrate that our framework achieves much better performance than state-of-the-art methods and exhibits a superior capacity for detecting rumors at early stages.

\end{abstract}

\section{Introduction}
With the proliferation of social media such as Twitter and Weibo, the emergence of breaking events provides opportunities for the spread of rumors, which is difficult to be identified due to limited domain expertise and relevant data. For instance, along with the unprecedented COVID-19 pandemic, a false rumor claimed that ``everyone who gets the vaccine will die or suffer from auto-immune diseases"\footnote{\url{https://www.bbc.com/news/uk-wales-58103604}} was translated into many languages and spread at lightning speed on social media, which seriously confuses the public and destroys the achievements of epidemic prevention in related countries or regions of the world. 
Although some recent works focus on collecting microblog posts corresponding to COVID-19~\cite{chen2020covid, zarei2020first, alqurashi2020large}, existing rumor detection methods perform poorly without a large-scale qualified training corpus, i.e., in a low-resource scenario~\cite{hedderich2021survey}. 
Thus there is an urgent need to develop automatic approaches to identify rumors in such low-resource domains especially amid breaking events. 

{\setlength{\abovecaptionskip}{-0.1cm}
\setlength{\belowcaptionskip}{-0.1cm}
\begin{figure*}[ht]
\centering
\subfigure[TWITTER (Rumor)]{
\begin{minipage}[t]{0.33\linewidth}
\centering
\scalebox{0.85}{\includegraphics[width=6cm]{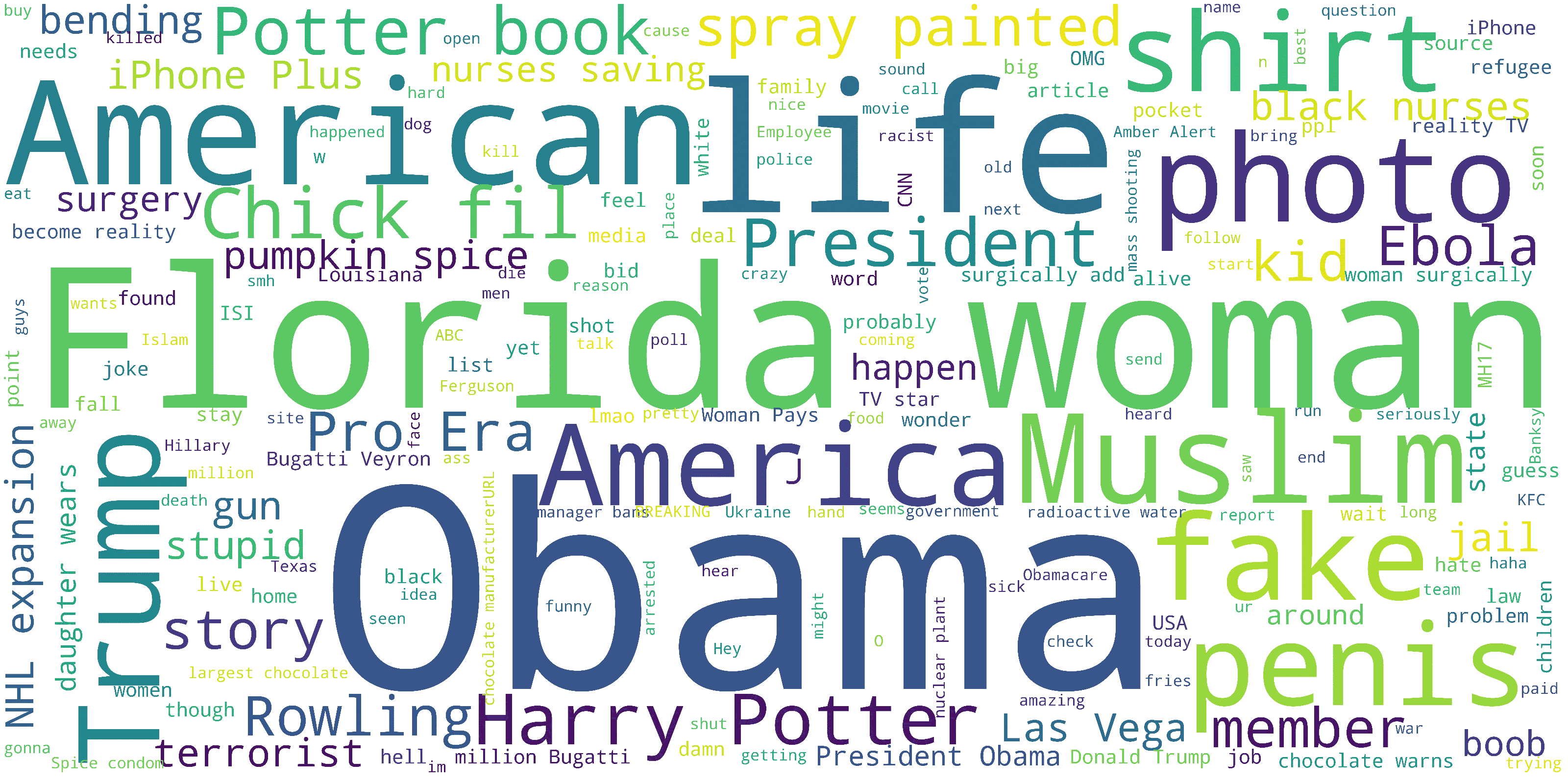}}
\label{fig:motivation_a}
\end{minipage}%
}%
\subfigure[Twitter-COVID19 (Rumor)]{
\begin{minipage}[t]{0.33\linewidth}
\centering
\scalebox{0.85}{\includegraphics[width=6cm]{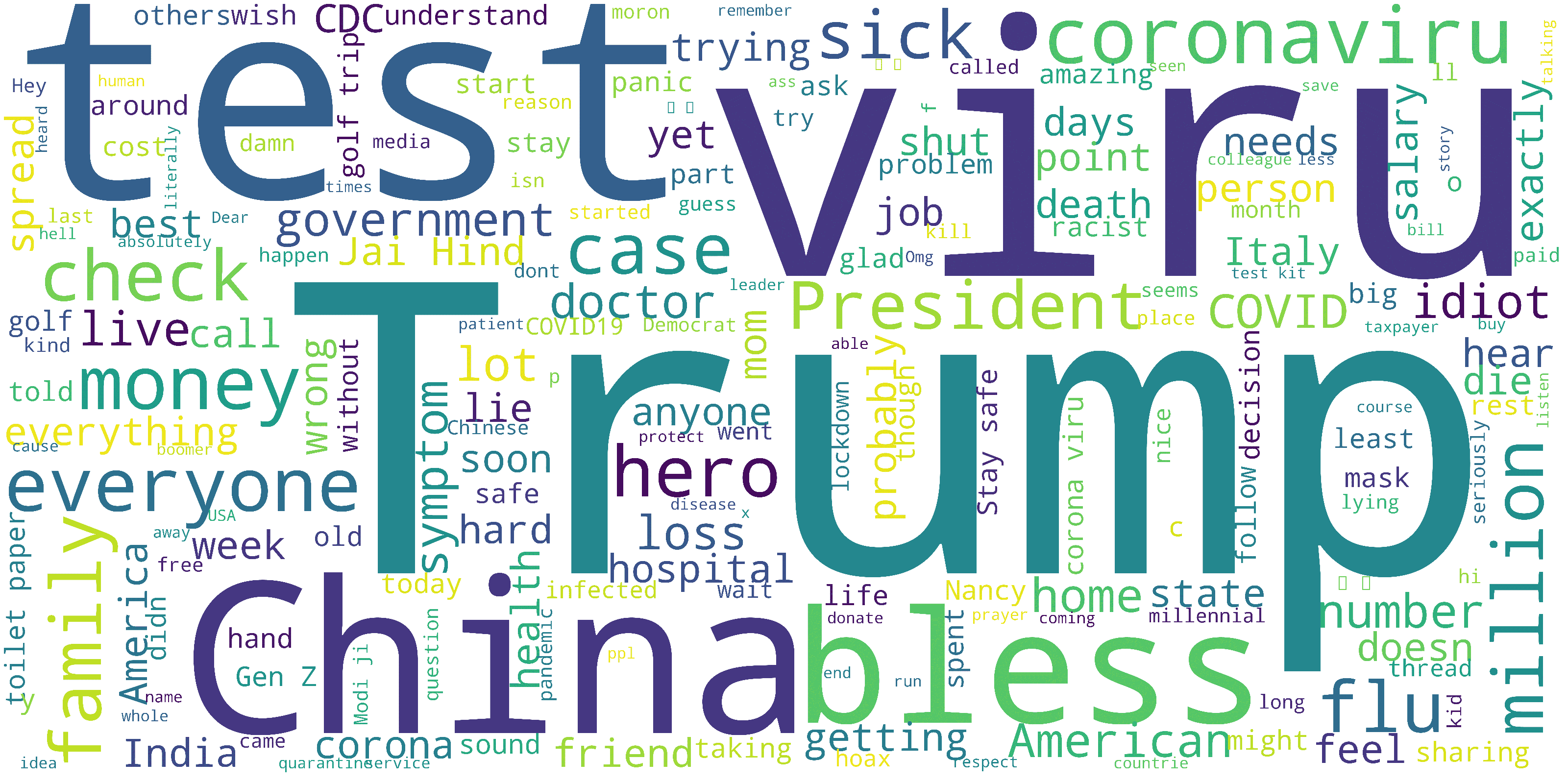}}
\label{fig:motivation_b}
\end{minipage}%
}%
\subfigure[Weibo-COVID19 (Rumor)]{
\begin{minipage}[t]{0.33\linewidth}
\centering
\scalebox{0.85}{\includegraphics[width=6cm]{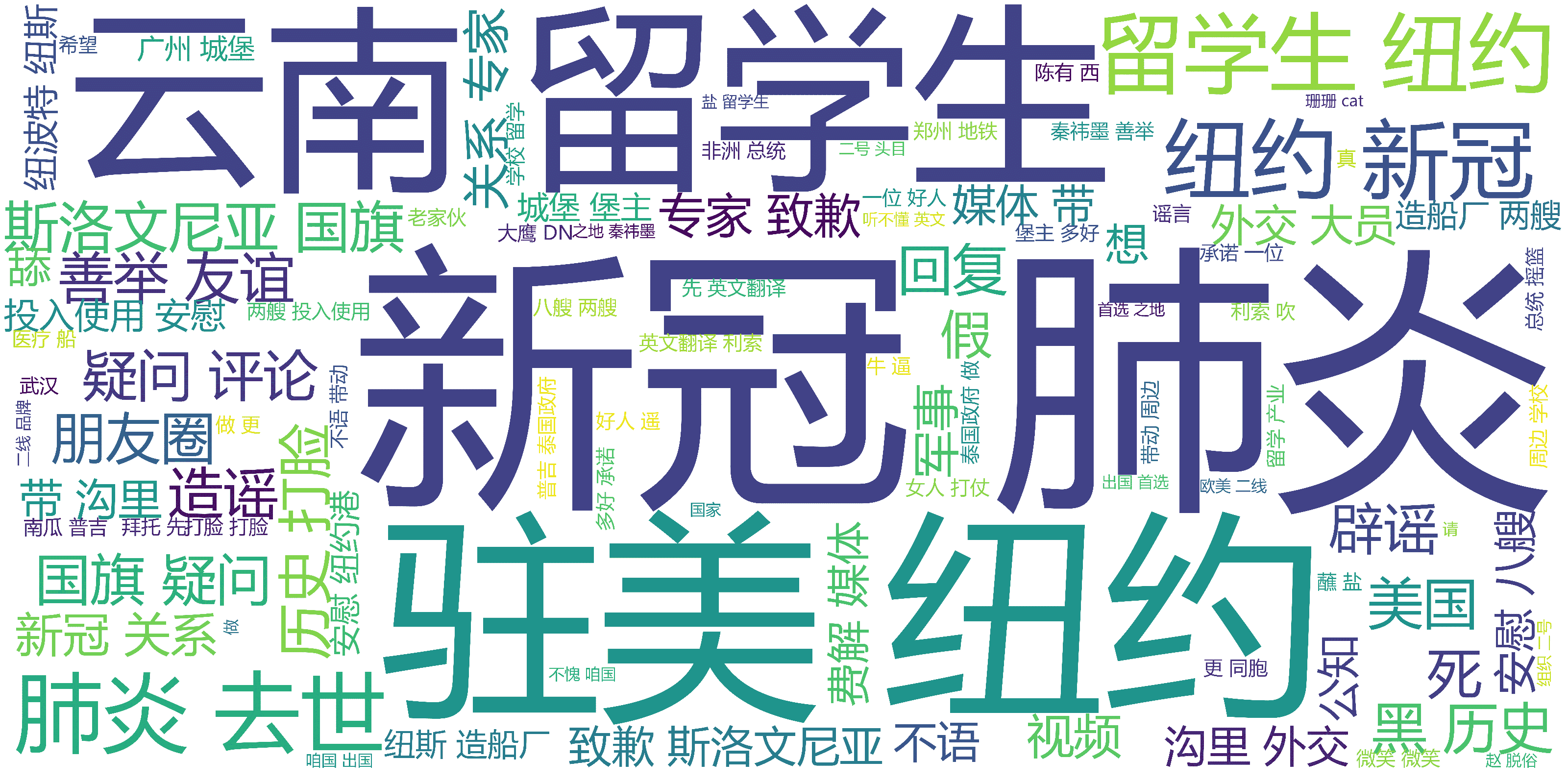}}
\label{fig:motivation_c}
\end{minipage}%
}%
\\
\subfigure[TWITTER (Non-rumor)]{
\begin{minipage}[t]{0.33\linewidth}
\centering
\scalebox{0.85}{\includegraphics[width=6cm]{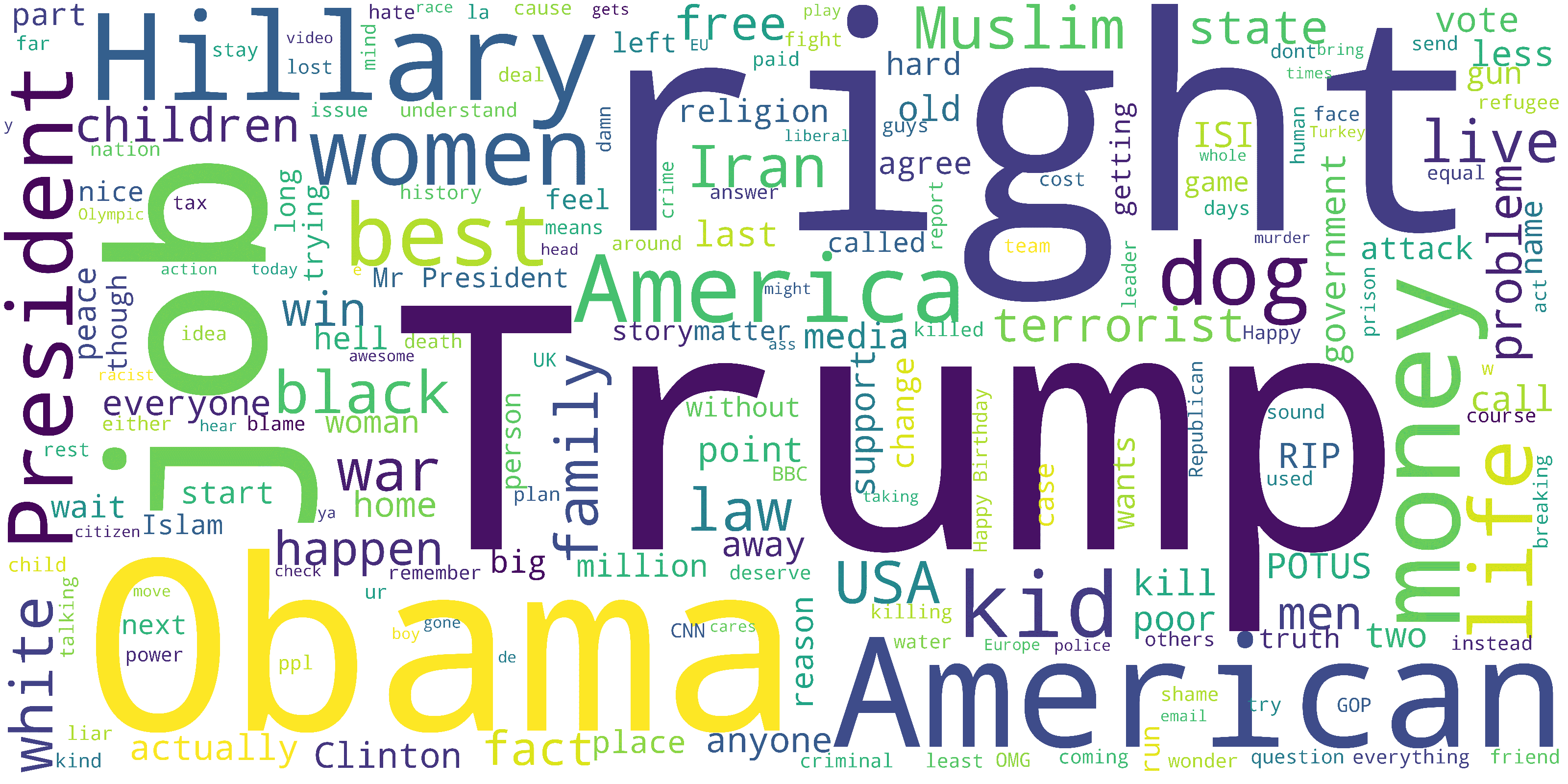}}
\label{fig:motivation_d}
\end{minipage}%
}%
\subfigure[Twitter-COVID19 (Non-rumor)]{
\begin{minipage}[t]{0.33\linewidth}
\centering
\scalebox{0.85}{\includegraphics[width=6cm]{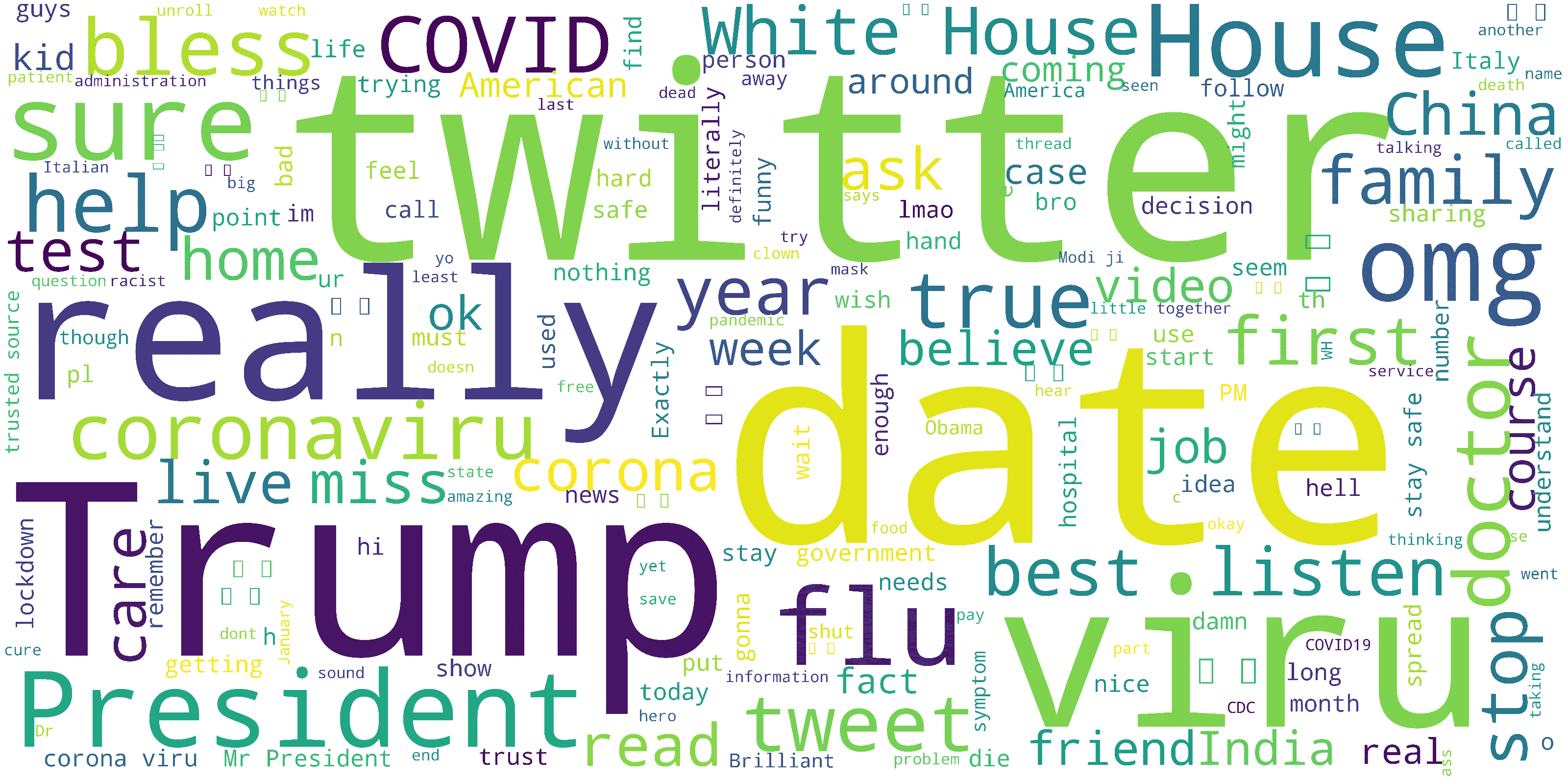}}
\label{fig:motivation_e}
\end{minipage}%
}%
\subfigure[Weibo-COVID19 (Non-rumor)]{
\begin{minipage}[t]{0.33\linewidth}
\centering
\scalebox{0.85}{\includegraphics[width=6cm]{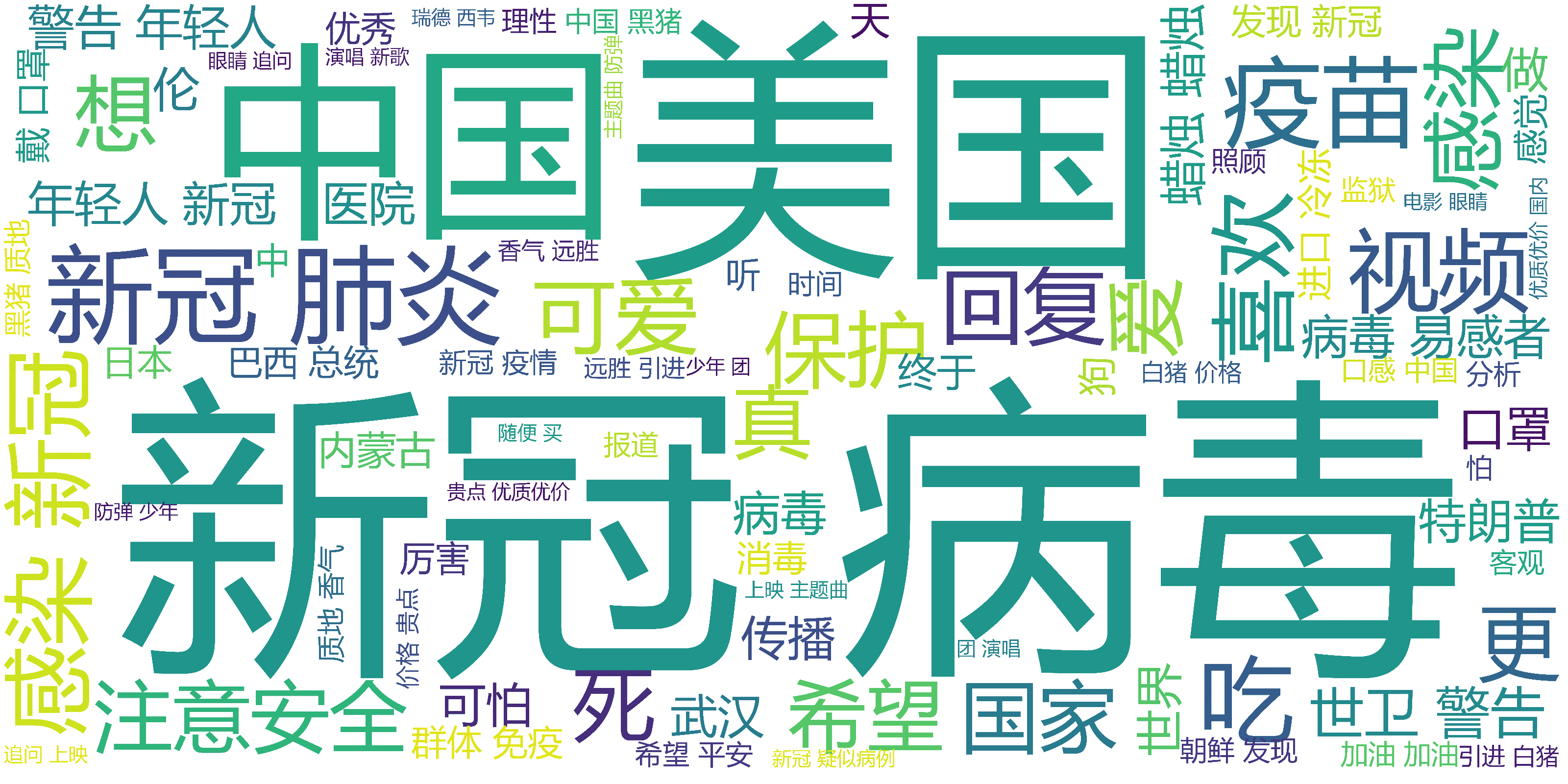}}
\label{fig:motivation_f}
\end{minipage}%
}%
\centering
\caption{Word clouds of rumor and non-rumor data generated from TWITTER, Twitter-COVID19, and Weibo-COVID19 datasets, where the size of terms corresponds to the word frequency. Both TWITTER and Twitter-COVID19 are presented in English while Weibo-COVID19 in Chinese.}
\label{fig:motivation}
\vspace{-0.2cm}
\end{figure*}}

Social psychology literature defines a rumor as a story or a statement whose truth value is unverified or deliberately false~\cite{allport1947psychology}. 
Recently, techniques using deep neural networks (DNNs)~\cite{ma2018rumor, khoo2020interpretable, bian2020rumor} have achieved promising results for detecting rumors on microblogging websites by learning rumor-indicative features from sizeable rumor corpus with veracity annotation. However, such DNN-based approaches are purely data-driven and have a major limitation on detecting emerging events concerning about low-resource domains, i.e., the distinctive topic coverage and word distribution~\cite{silva2021embracing} required for detecting low-resource rumors are often not covered by the public benchmarks~\cite{zubiaga2016learning, ma2016detecting, ma2017detect}. 
On another hand, for rumors propagated in different languages, existing monolingual approaches are not applicable since there are even no sufficient open domain data for model training in the target language. 


In this paper, we assume that the close correlations between the well-resourced rumor data and the low-resourced could break the barriers of domain and language, substantially boosting low-resource rumor detection within a more general framework. Taking the breaking event COVID-19 as an example, we collect corresponding rumorous and non-rumorous claims with propagation threads from Twitter and Sina Weibo which are the most popular microblogging websites in English and Chinese, respectively. Figure~\ref{fig:motivation} illustrates the word clouds of rumor and non-rumor data from an open domain benchmark (i.e., TWITTER ~\cite{ma2017detect}) and two COVID-19 datasets (i.e., Twitter-COVID19 and Weibo-COVID19). It can be seen that both TWITTER and Twitter-COVID19 contain denial opinions towards rumors, e.g., ``fake", ``joke", ``stupid" in Figure~\ref{fig:motivation_a} and ``wrong symptom", ``exactly sick", ``health panic" in Figure~\ref{fig:motivation_b}. In contrast, supportive opinions towards non-rumors can be drawn from Figure~\ref{fig:motivation_d}--\ref{fig:motivation_e}. 
Moreover, considering that COVID-19 is a global disease, massive misinformation could be widely propagated in different languages such as Arabic~\cite{alam2020fighting}, Indic~\cite{kar2020no}, English~\cite{cui2020coaid} and Chinese~\cite{hu2020weibo}. Similar identical patterns can be observed in Chinese on Weibo from Figure~\ref{fig:motivation_c} and Figure~\ref{fig:motivation_f}.
Although the COVID-19 data tend to use expertise words or language-related slang, we argue that aligning the representation space of identical rumor-indicative patterns of different domains and/or languages could adapt the features captured from well-resourced data to that of the low-resourced.

To this end, inspired by contrastive learning~\cite{he2020momentum, chen2020simple, chen2020big}, we propose an Adversarial Contrastive Learning approach for low-resource rumor detection (ACLR), to encourage effective alignment of rumor-indicative features in the well-resourced and low-resource data. More specifically, we first transform each microblog post into a language-independent vector by semantically aligning the source and target language in a shared vector space. 
As the diffusion of rumors generally follows a propagation tree that provides valuable clues on how a claim is transmitted~\cite{ma2018rumor}, we thus resort to a structure-based neural network~\cite{bian2020rumor} to catch informative patterns. Then,
we propose a novel supervised contrastive learning paradigm to minimize the intra-class variance of source and target instances with same veracity, and maximize inter-class variance of instances with different veracity. 
To further enhance the feature adaption of contrastive learning, we exploit adversarial attacks~\cite{ kurakin2016adversarial} to plenish noise to the original event-level representation by computing adversarial worst-case perturbations, forcing the model to learn non-trivial but effective features. 
Extensive experiments conducted on two real-world low-resource datasets confirm that (1) our model yields outstanding performances for detecting low-resource rumors over the state-of-the-art baselines with a large margin; and (2) our method performs particularly well on early rumor detection which is crucial for timely intervention and debunking especially for breaking events. 
%
The main contributions of this paper are of three-fold:
\begin{itemize}
\item To our best knowledge, we are the first to present a radically novel adversarial contrastive learning framework to study the low-resource rumor detection on social media\footnote{Our resources will be available at \url{https://github.com/DanielLin97/ACLR4RUMOR-NAACL2022}.}.
\item We propose supervised contrastive learning for structural feature adaption between different domains and languages, with adversarial attacks employed to enhance the diversity of low-resource data for contrastive paradigm.
\item We constructed two low-resource microblog datasets corresponding to COVID-19 with propagation tree structure, respectively gathered from English tweets and Chinese microblog posts. Experimental results show that our model achieves superior performance for both rumor classification and early detection tasks under low-resource settings.
\end{itemize}

\section{Related Work}
Pioneer studies for automatic rumor detection focus on learning a supervised classifier utilizing features crafted from post contents, user profiles, and propagation patterns~\cite{castillo2011information, yang2012automatic, liu2015real}. Subsequent studies then propose new features such as those representing rumor diffusion and cascades \cite{kwon2013prominent, friggeri2014rumor, hannak2014get}. \citet{zhao2015enquiring} alleviate the engineering effort by using a set of regular expressions to find questing and denying tweets. DNN-based models such as recurrent neural networks~\cite{ma2016detecting}, convolutional neural networks~\cite{yu2017convolutional}, and attention mechanism~\cite{guo2018rumor} are then employed to learn the features from the stream of social media posts. However, these approaches simply model the post structure as a sequence while ignoring the complex propagation structure. 

To extract useful clues jointly from content semantics and propagation structures, some approaches propose kernel-learning models~\cite{wu2015false, ma2017detect} to make a comparison between propagation trees. Tree-structured recursive neural networks (RvNN)~\cite{ma2018rumor} and transformer-based models~\cite{khoo2020interpretable, ma2020debunking} are proposed to generate the representation of each post along a propagation tree guided by the tree structure. 
More recently, graph neural networks~\cite{bian2020rumor,lin2021rumor} have been exploited to encode the conversation thread for higher-level representations. 
However, such data-driven approaches fail to detect rumors in low-resource regimes~\cite{janicka2019cross} because they often require sizeable training data which is not available for low-resource domains and/or languages. In this paper, we propose a novel framework 
to adapt existing models with the effective propagation structure for detecting rumors from different domains and/or languages. 

To facilitate related fact-checking tasks in low-resource settings, domain adaption techniques are utilized to detect fake news~\cite{wang2018eann, yuan2021improving, zhang2020bdann, silva2021embracing} by learning features from multi-modal data such as texts and images. 
\citet{lee2021towards} proposed a simple way of leveraging the perplexity score obtained from pre-trained language models (LMs) for the few-shot fact-checking task. 
Different from these works of adaption on multi-modal data and transfer learning of LMs, we focus on language and domain adaptation to detect rumors from low-resource microblog posts corresponding to breaking events. 


Contrastive learning (CL) aims to enhance representation learning by maximizing the agreement among the same types of instances and distinguishing from the others with different types~\cite{wang2020understanding}.  
In recent years, CL has achieved great success in unsupervised visual representation learning~\cite{chen2020simple, he2020momentum, chen2020big}. 
Besides computer vision, recent studies suggest that CL is promising in the semantic textual similarity~\cite{gao2021simcse, yan2021consert}, stance detection~\cite{mohtarami2019contrastive}, short text clustering~\cite{zhang2021supporting}, unknown intent detection~\cite{ lin2021boosting}, and abstractive summarization~\cite{liu2021simcls}, etc. However, the above CL frameworks are specifically proposed to augment unstructured textual data such as sentence and document, which are not suitable for the low-resource rumor detection task considering claims together with more complex propagation structures of community response. 

\begin{figure*}[t]
\centering
\scalebox{0.75}{\includegraphics[width=20cm]{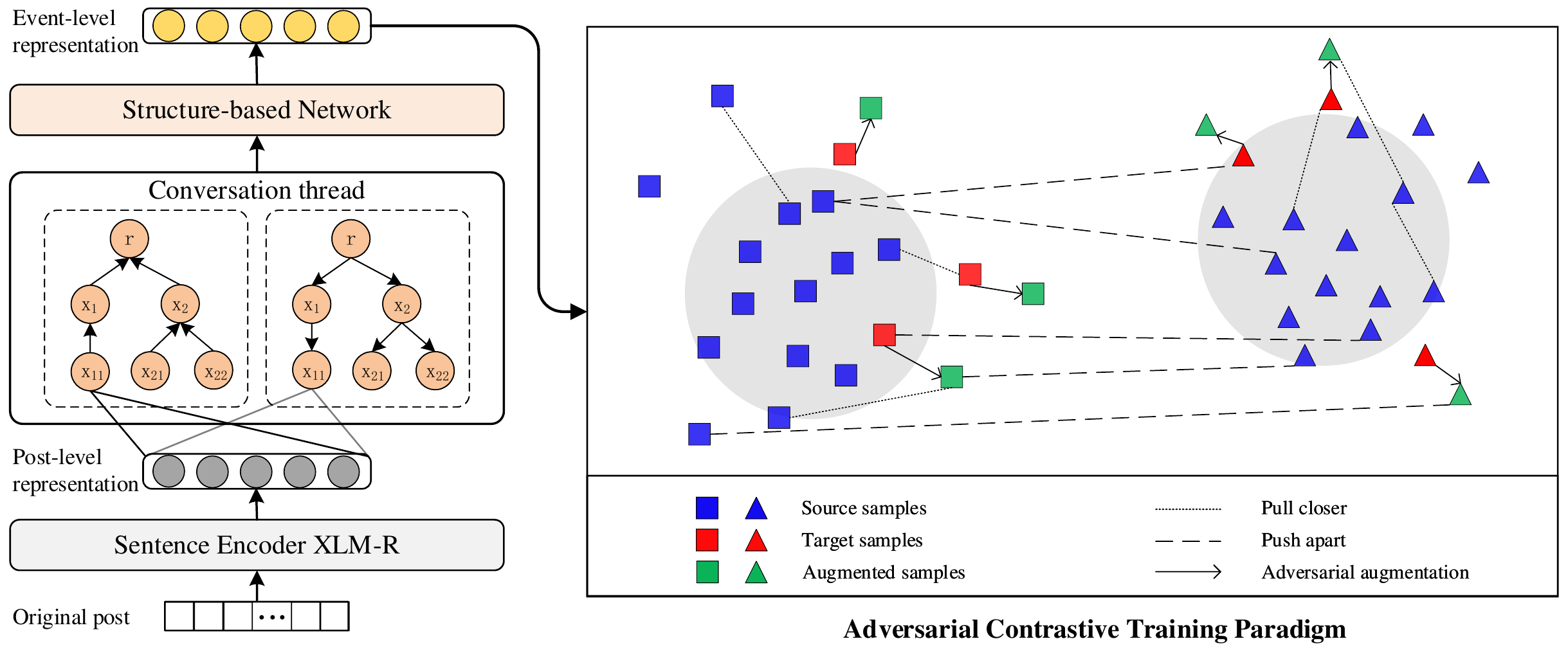}}
\caption{The overall architecture of our proposed method. For source and small target training data, we first obtain post-level representations after cross-lingual sentence encoding, then train the structure-based network with the adversarial contrastive objective. For target test data, we extract the event-level representations to detect rumors. }
\label{fig:method}
\vspace{-0.5cm}
\end{figure*}

\section{Problem Statement}
In this work, we define the low-resource rumor detection task as: Given a well-resourced dataset as source, classify each event in the target low-resource dataset as a rumor or not, where the source and target data are from different domains and/or languages. Specifically, we define a well-resourced source dataset for training as a set of events $\mathcal{D}_s = \{C_1^s, C_2^s, \cdots, C_{M}^s\}$, where $M$ is the number of source events. Each event $C^s=(y,c,\mathcal{T}(c))$ is a tuple representing a given claim $c$ which is associated with a veracity label $y \in \{\text{rumor}, \text{non-rumor}\}$, and ideally all its relevant responsive microblog post in chronological order, i.e., $\mathcal{T}(c) = \{c, x_{1}^s,x_{2}^s,\cdots,x_{|C|}^{s}\}$\footnote{$c$ is equivalent to $x_0^s$.}, where 
$|C|$ is the number of responsive tweets in the conversation thread. 
For the target dataset with low-resource domains and/or languages, we consider a much smaller dataset for training $\mathcal{D}_t = \{C_1^t, C_2^t, \cdots, C_{N}^t\}$, where $N (N \ll M)$ is the number of target events and each $C^t=(y,c',\mathcal{T}(c'))$ 
has the similar composition structure of the source dataset. 

We formulate the task of low-resource rumor detection as a supervised classification problem that trains a domain/language-agnostic classifier $f(\cdot)$ adapting the features learned from source datasets to that of the target events, that is, $f(C^t| \mathcal{D}_s) \rightarrow y$. Note that although the tweets are notated sequentially, there are connections among them based on their responsive relationships. So most previous works represent the conversation thread as a tree structure~\cite{ma2018rumor, bian2020rumor}.

\section{Our Approach}
In this section, we introduce our adversarial contrastive learning framework to adapt the features captured from the well-resourced data to detect rumors from low-resource events, which considers cross-lingual and cross-domain alignment. Figure~\ref{fig:method} illustrates an overview of our proposed model, which will be depicted in the following subsections.

\subsection{Cross-lingual Sentence Encoder}
Given a post in an event that could be either from source or target data, to map it into a shared semantic space where the source and target languages are semantically aligned, we utilize XLM-RoBERTa~\cite{conneau2019unsupervised} (XLM-R) to model the context interactions among tokens in the sequence for the sentence-level representation:
{\setlength{\abovedisplayskip}{0.1cm}
\setlength{\belowdisplayskip}{0.1cm}
\begin{equation}\label{eq:1}
    \bar{x} = \textit{XLM-R}(\textbf{x})
\end{equation}
}where $\textbf{x}$ is the original post, and we obtain the post-level representation $\bar{x}$ using the output state of the <$s$> token in XLM-R. We thus denote the representation of posts in the source event $C^s$ and the target event $C^t$ as a matrix $X^s$ and $X^t$ respectively: $X^* = {[\bar{x}_0^*, \bar{x}_1^*,\bar{x}_2^*,...,\bar{x}_{|X^*|-1}^*]}^{\top}; *\in\{s,t\}$, where $X^s \in \mathbb{R}^{m\times d}$ and $X^t \in \mathbb{R}^{n \times d}$, $d$ is the dimension of the output state of the sentence encoder.

\subsection{Propagation Structure Representation}
\label{BiGCN}
On top of the sentence encoder, we represent the propagation of each claim with the graph convolutional network (GCN)~\cite{kipf2016semi}, which achieves state-of-the-art performance on capturing both structural and semantic information for rumor classification~\cite{bian2020rumor}. It is worth noting that the choice of propagation structure representation is orthogonal to our proposed framework that can be easily replaced with any existing structure-based models without any other change to our supervised contrastive learning architecture. 

Given an event and its initialized embedding matrix $C^*, X^*; * \in \{s,t\}$, We model the conversation thread of the event as a tree structure $\mathcal{T}=\langle V,E \rangle$, where $V$ consists of the event claim and all its relevant responsive posts as nodes and $E$ refers to a set of directed edges corresponding to the response relation among the nodes in $V$. Inspired by \citet{ma2018rumor}, here we consider two different propagation trees with distinct edge directions: (1) \textit{Top-Down tree} where the edge follows the direction of information diffusion. (2) \textit{Bottom-Up tree} where the responsive nodes point to their responded nodes, similar to a citation network.

\textbf{Top-Down GCN.} We treat the Top-Down tree structure as a graph and transform the edge $E$ into an adjacency matrix $\textbf{A} \in {\{0,1\}}^{|{V}| \times |{V}|}$, where $\textbf{A}_{i,j} = 1$ if $\textbf{x}_{i}$ has a response to $\textbf{x}_{j}$ or $i=j$, else $\textbf{A}_{i,j} = 0$. Then we utilize a layer-wise propagation rule to update the node vector at the $l$-th layer: 
{\setlength{\abovedisplayskip}{0.1cm}
\setlength{\belowdisplayskip}{0.1cm}
\begin{equation}
    H^{(l+1)} = ReLU(\hat{\textbf{A}} \cdot H^{(l)} \cdot W^{(l)})
    \label{equ:GCN}
\end{equation}} where $\hat{\textbf{A}} = \textbf{D}^{-1/2} \textbf{A} \textbf{D}^{-1/2}$ is the symmetric normalized adjacency matrix, $\textbf{D}$ denotes the degree matrix of $\textbf{A}$. $W^{(l)} \in \mathbb{R}^{d^{(l)} \times d^{(l+1)}}$ is
a layer-specific trainable transformation matrix. 
$H^{(0)}$ is initialized as $X^*$. For a GCN model with $L$-layers, we obtain the final node representation $H_{TD}$ w.r.t $H^{(L)}$. 

\textbf{Bottom-Up GCN.} We also leverage the structure of Bottom-Up tree to encode the informative posts. 
Similar to Top-Down GCN, we update the hidden representation of nodes in the same manner as Eq.~\ref{equ:GCN} and finally get the output node states $H_{BU}$ at the $L$-th graph convolutional layer.

\textbf{The Overall Model.} Finally, we concatenate $H_{TD}$ and $H_{BU}$ 
via mean-pooling to jointly capture the opinions expressed in both Top-Down and Bottom-Up trees:
{\setlength{\abovedisplayskip}{0.1cm}
\setlength{\belowdisplayskip}{0.1cm}
\begin{equation}\label{eq:3}
    o = \emph{mean-pooling}([H_{TD};H_{BU}])
\end{equation}} where $o \in \mathbb{R}^{2d^{(L)}}$ is the event-level representation of the entire propagation thread, $d^{(L)}$ is the output dimension of GCN and $[{\cdot} ; {\cdot}]$ means concatenation.

\subsection{Contrastive Training}
To align the representation space of rumor-indicative signals from different domains and languages, we present a novel training paradigm to exploit the labeled data including rich sourced data and small-scaled target data to adapt our model on target domains and languages. The core idea is to make the representations of source and target events from the same class closer while keeping representations from different classes far away.

Given an event $C_i^s$ from the source data, we firstly obtain the language-agnostic encoding for all the involved posts (see Eq.~\ref{eq:1}) as well as 
the propagation structure representation $o_i^s$ (see Eq.~\ref{eq:3}) which is then fed into a \textit{softmax} function to make rumor predictions. 
Then, we learn to minimize the cross-entropy loss between the prediction and the ground-truth label $y_i^s$: 
{\setlength{\abovedisplayskip}{0.1cm}
\setlength{\belowdisplayskip}{0.1cm}
\begin{equation}
    \mathcal{L}_{CE}^s = -\frac{1}{N^s} \sum\limits_{i=1}^{N^s} log(p_i)
    \label{equ:ce}
\end{equation}} where $N^s$ is the total number of source examples in the batch, $p_i$ is the probability of correct prediction. 
To make rumor representation in the source events be more dicriminative, we propose a supervised contrastive learning objective to cluster the same class and separate different classes of samples:
{\setlength{\abovedisplayskip}{0.1cm}
\setlength{\belowdisplayskip}{0.1cm}
\begin{equation}
\begin{split}
    \mathcal{L}_{SCL}^s=-\frac{1}{N^s}\sum\limits_{i=1}^{N^s} \frac{1}{N_{y_i^s}-1} \sum\limits_{j=1}^{N^s}\mathbbm{1}_{[i \ne j]}\mathbbm{1}_{[y_i^s = y_j^s]}\\
    log\frac{\emph{exp}(\text{sim}(o_i^s, o_j^s)/\tau)}{\sum\limits_{k=1}^{N^s} \mathbbm{1}_{[i \ne k]} \emph{exp}(\text{sim}(o_i^s, o_k^s)/\tau)}
\end{split}
\end{equation}} where $N_{y_i^s}$ is the number of source examples with the same label $y_i^s$ in the event $C_i^s$, and $\mathbbm{1}$ is the indicator. $\text{sim}(\cdot)$ denotes the cosine similarity function and $\tau$ controls the temperature. 

For an event $C_i^t$ from the target data, we also compute the classification loss $\mathcal{L}_{CE}^t$ in the same manner as Eq.~\ref{equ:ce}. Although we projected the source and target languages into the same semantic space after sentence encoding, rumor detection not only relies on post-level features, but also on event-level contextual features. Without constraints, the structure-based network can only extract event-level features for all samples based on their final classification signals while these features may not be critical to the target domain and language. We make full use of the minor labels in the low-resource rumor data by parameterizing our model according to the contrastive objective between the source and target instances in the event-level representation space:
{\setlength{\abovedisplayskip}{0.1cm}
\setlength{\belowdisplayskip}{0.1cm}
\begin{equation}
\begin{split}
  \mathcal{L}_{SCL}^t = -\frac{1}{N^t}\sum\limits_{i=1}^{N^t} \frac{1}{N_{y_i^t}} \sum\limits_{j=1}^{N^s}\mathbbm{1}_{[y_i^t = y_j^s]}\\
    log\frac{\emph{exp}(\text{sim}(o_i^t, o_j^s)/\tau)}{\sum\limits_{k=1}^{N^s} \emph{exp}(\text{sim}(o_i^t, o_k^s)/\tau)}
 \end{split}
\end{equation}} where $N^t$ is the total number of target examples in the batch and $N_{y_i^t}$ is the number of source examples with the same label $y_i^t$ in the event $C_i^t$. 
As a result, we project the source and target samples belonging to the same class closer than that of different categories, 
for feature alignment with minor annotation at the target domain and language. 

\subsection{Adversarial Data Augmentation}
Data augmentation techniques were successfully utilized to enhance contrastive learning models~\cite{chen2020simple}. Some simple augmentation strategies are designed based on handcrafted features or rules, but they are not efficient and suitable for the propagation tree structures in rumor detection task. 
In this section, we introduce adversarial attacks to generate pseudo target samples at the event-level latent space to increase the diversity of views for model robustness in the contrastive learning manner. Specifically, we apply Fast Gradient Value~\cite{miyato2016adversarial,vedula2020open} to approximate a worst-case perturbation as a noise vector of the event-level representation:
{\setlength{\abovedisplayskip}{0.1cm}
\setlength{\belowdisplayskip}{0.1cm}
\begin{equation}
    \tilde{\bm{o}}_{noise}^t= \epsilon \frac{g}{||g||}; \text{where} \ \ g = \nabla_{o^t} \mathcal{L}_{CE}^t
\end{equation}} 
where the gradient is the first-order differential of the classification loss $\mathcal{L}_{CE}^t$ for a target sample, 
i.e., the direction that rapidly increases the classification loss. We perform normalization and use a small $\epsilon$ to ensure the approximate is reasonable. Finally, we can obtain the pseudo augmented sample $o_{adv}^t = o^t +\tilde{\bm{o}}_{noise}^t$ in the latent space to enhance our model.

\begin{algorithm}[t]
\small
  \caption{\textbf{Adversarial Contrastive Learning}}
  \label{algorithm}
  \begin{algorithmic}[1]
    \Require
      A small set of events $C_i^t$ in the target domain and language; A set of events $C_i^s$ in the source domain and language.
    \Ensure
      Assign rumor labels $y$ to given unlabeled target data.
	\State \textbf{for} each mini-batch $N^t$ of the target events $C_i^t$ \textbf{do}:
    \State \quad \textbf{for} each mini-batch $N^s$ of the source events $C_i^s$ \textbf{do}:
    \State \quad \quad Pass $C_i^*$ to the sentence encoder and then structure-based network to obtain its event-level feature $o_i^*$, where $* \in \{s, t\}$. 
    \State \quad \quad Compute the classification loss $\mathcal{L}_{CE}^*$ for source and target data, respectively.
    \State \quad \quad Adversarial augmentation for target data and update $\mathcal{L}_{CE}^t$.
    \State \quad \quad Compute the supervised contrastive loss $\mathcal{L}_{SCL}^*$.
    \State \quad \quad Compute the joint loss $\mathcal{L}^*$ as Eq.~\ref{joint_loss}.
    \State \quad \quad Jointly optimize all parameters of the model using the average loss $\mathcal{L} = \text{mean}({\mathcal{L}^s}+{\mathcal{L}^t})$.
  \end{algorithmic}
\end{algorithm}

\subsection{Model Training}
We jointly train the model with the cross-entropy and supervised contrastive objectives:
{\setlength{\abovedisplayskip}{0.1cm}
\setlength{\belowdisplayskip}{0.1cm}
\begin{equation}
    \mathcal{L}^* = (1-\alpha)\mathcal{L}_{CE}^* + \alpha  \mathcal{L}_{SCL}^*; * \in \{s, t\}
    \label{joint_loss}
\end{equation}} 
where $\alpha$ is a trade-off parameter, which is set to 0.5 in our experiments. Algorithm~\ref{algorithm} presents the training process of our approach. 
We set the number $L$ of the graph convolutional layer as 2, the temperature $\tau$ as 0.1, and the adversarial perturbation norm $\epsilon$ as 1.5. Parameters are updated through back-propagation~\cite{collobert2011natural} with the Adam optimizer~\cite{loshchilov2018decoupled}. The learning rate is initialized as 0.0001, and the dropout rate is 0.2. Early stopping~\cite{yao2007early} is applied to avoid overfitting.

\section{Experiments}
\subsection{Datasets}
To our knowledge, there are no public benchmarks available for detecting low-resource rumors with propagation tree structure in tweets. In this paper, we consider a breaking event COVID-19 as a low-resource domain and collect relevant rumors and non-rumors respectively from Twitter in English and Sina Weibo in Chinese. 
For Twitter-COVID19, we resort to a COVID-19 rumor dataset~\cite{kar2020no} which only contains textual claims without propagation thread. We extend each claim by collecting its propagation threads via Twitter academic API with a twarc2 package\footnote{\url{https://twarc-project.readthedocs.io/en/latest/twarc2_en_us/}}. For Weibo-COVID19, similar to \citet{ma2016detecting}, a set of related rumorous claims are gathered from the Sina community management center\footnote{\url{https://service.account.weibo.com/}} and non-rumorous claims by randomly filtering out the posts that are not reported as rumors. Then Weibo API is utilized to collect all the repost/reply messages towards each claim (see Appendix for the dataset statistics). 

\begin{table*}[t]\small
\centering
\begin{center}
\resizebox{0.97\textwidth}{!}{
\begin{tabular}{l||cc|cc||cccc}
\hline
Target (Source)        & \multicolumn{4}{c|}{Weibo-COVID19 (TWITTER)}                                      & \multicolumn{4}{c}{Twitter-COVID19 (WEIBO)}                                                            \\ \hline
\multirow{2}{*}{Model} 
                       & \multirow{2}{*}{Acc.} & \multirow{2}{*}{Mac-$\emph{F}_1$ } & Rumor          & Non-rumor      & \multirow{2}{*}{Acc.} & \multicolumn{1}{c|}{\multirow{2}{*}{Mac-$\emph{F}_1$ }} & Rumor          & Non-rumor      \\ \cline{4-5} \cline{8-9} 
                       &                       &                         & $\emph{F}_1$              & $\emph{F}_1$              &                       & \multicolumn{1}{c|}{}                        & $\emph{F}_1$              & $\emph{F}_1$              \\ \hline \hline
CNN              & 0.445                 & 0.402                   & 0.476          & 0.328              & 0.498                 & \multicolumn{1}{c|}{0.389}                   & 0.528          & 0.249              \\
RNN          & 0.463                 & 0.414                   & 0.498              & 0.329          & 0.510                 & \multicolumn{1}{c|}{0.388}                   & 0.533              & 0.243          \\ \hline
RvNN                   & 0.514                 & 0.482                   & 0.538          & 0.426          & 0.540                 & \multicolumn{1}{c|}{0.391}                   & 0.534          & 0.247          \\
PLAN                   & 0.532                 & 0.496                   & 0.578          & 0.414          & 0.573                 & \multicolumn{1}{c|}{0.423}                   & 0.549          & 0.298          \\
BiGCN                  & 0.569                 & 0.508                   & 0.586          & 0.429          & 0.616                 & \multicolumn{1}{c|}{0.415}                   & 0.577          & 0.252          \\ \hline
DANN-RvNN              & 0.583                 & 0.498                   & 0.591          & 0.405          & 0.577                 & \multicolumn{1}{c|}{0.482}                   & 0.648          & 0.317          \\
DANN-PLAN              & 0.601                 & 0.507                   & 0.606          & 0.409          & 0.593                 & \multicolumn{1}{c|}{0.471}                   & 0.574          & 0.369          \\
DANN-BiGCN             & 0.629                 & 0.561                   & 0.616          & 0.506          & 0.618                 & \multicolumn{1}{c|}{0.510}                   & 0.676          & 0.344          \\ \hline
ACLR-RvNN               & 0.778                 & 0.716                   & 0.843          & 0.589          & 0.653                 & \multicolumn{1}{c|}{0.616}                   & 0.710          & 0.521          \\
ACLR-PLAN               & 0.824                 & 0.769                   & 0.842          & 0.696          & 0.709                 & \multicolumn{1}{c|}{0.648}                   & 0.752          & 0.544          \\
ACLR-BiGCN              & \textbf{0.873}        & \textbf{0.861}          & \textbf{0.896} & \textbf{0.827} & \textbf{0.765}        & \multicolumn{1}{c|}{\textbf{0.686}}          & \textbf{0.766} & \textbf{0.605} \\ \hline
\end{tabular}}
\end{center}
\caption{Rumor detection results on the target test datasets.}
\vspace{-0.2cm}
\label{tab:main_results}
\end{table*}

\subsection{Experimental Setup}
We compare our model and several state-of-the-art baseline methods described below. 1) \textbf{CNN}: A CNN-based model for misinformation identification~\cite{yu2017convolutional} by framing the relevant posts as a fixed-length sequence; 2) \textbf{RNN}: A RNN-based rumor detection model~\cite{ma2016detecting} with GRU for feature learning of relevant posts over time; 3) \textbf{RvNN}: A rumor detection approach based on tree-structured recursive neural networks \cite{ma2018rumor} that learn rumor representations guided by the propagation structure; 4) \textbf{PLAN}: A transformer-based model \cite{khoo2020interpretable} for rumor detection to capture long-distance interactions between any pair of involved tweets; 5) \textbf{BiGCN}: A GCN-based model~\cite{bian2020rumor} based on directed conversation trees to learn higher-level representations (see Section~\ref{BiGCN}); 6) \textbf{DANN-*}: We employ and extend an existing few-shot learning technique, domain-adversarial neural network~\cite{ganin2016domain}, based on the structure-based model where * could be RvNN, PLAN, and BiGCN; 7) \textbf{ACLR-*}: our proposed adversarial contrastive learning framework on top of RvNN, PLAN, or BiGCN.

In this work, we consider the most challenging setting: to detect events (i.e., target) from a low-resource domain meanwhile in a cross-lingual regime. 
Note that although English and Chinese in our datasets are not minority languages, the target domain and/or languages can be easily replaced without any change to our ACLR framework. Specifically, we use the well-resourced TWITTER~\cite{ma2017detect} (or WEIBO~\cite{ma2016detecting}) datasets as the source data, and Weibo-COVID19 (or Twitter-COVID19) datasets as the target. 
We use accuracy and macro-averaged F1, as well as class-specific F1 scores as the evaluation metrics. We conduct 5-fold cross-validation on the target datasets 
(see more details in Appendix).

\subsection{Rumor Detection Performance}
Table \ref{tab:main_results} shows the performance of our proposed method versus all the compared methods on the Weibo-COVID19 and Twitter-COVID19 test sets with pre-determined training datasets. It is observed that the performances of the baselines in the first group are obviously poor due to ignoring intrinsic structural patterns. To make fair comparisons, all baselines are employed with the same cross-lingual sentence encoder of our framework as inputs. Other state-of-the-art baselines exploit the structural property of community wisdom on social media, 
which confirms the necessity of propagation structure representations in our framework. 

Among the structure-based baselines in the second group, due to the representation power of message-passing architectures and tree structures, PLAN and BiGCN outperform RvNN with only limited labeled target data for training. The third group shows the results for DANN-based methods. It improves the performance of structure-based baselines in general since it extracts cross-domain features between source and target datasets via generative adversarial nets~\cite{goodfellow2014generative}. Different from that, we use the adversarial attacks to improve the robustness of our proposed contrastive training paradigm, which explicitly encourages effective alignment of rumor-indicative features from different domains and languages. 

In contrast, our proposed ACLR-based approaches achieve superior performances among all their counterparts ranging from 21.8\% (13.4\%) to 30.0\% (17.7\%) in terms of Macro F1 score on Weibo-COVID19 (Twitter-COVID19) datasets, which suggests their strong judgment on low-resource rumors from different domains/languages. ACLR-BiGCN performs the best among the three ACLR-based methods by making full use of the structural property via graph modeling for conversation threads. This also justifies the good performance of DANN-BiGCN and BiGCN. The results also indicate that the adversarial contrastive learning framework can effectively transfer knowledge from the source to target data at the event level, and substantiate our method is model-agnostic for different structure-based networks. 

\subsection{Ablation Study}
We perform ablation studies based on our best-performed approach ACLR-BiGCN. As demonstrated in Table~\ref{ablation}, the first group shows the results for the backbone baseline BiGCN. We observe that the model performs best if pre-trained on source data and then fine-tuned on target training data (i.e., BiGCN(S,T)), compared with the poor performance when trained on either minor labeled target data only (i.e., BiGCN(T)) or well-resourced source data (i.e., BiGCN(S)). This suggests that our hypothesis of leveraging well-resourced source data to improve the low-resource rumor detection on target data is feasible. In the second group, the DANN-based model makes better use of the source data to extract domain-agnostic features, which further leads to performance improvement. Our proposed contrastive learning approach CLR without adversarial augmentation mechanism, has already achieved outstanding performance compared with other baselines, which illustrates its effectiveness on domain and language adaptation. We further notice that our ACLR-BiGCN consistently outperforms all baselines and improves the prediction performance of CLR-BiGCN, suggesting that training model together with adversarial augmentation on target data provide positive guidance for more accurate rumor predictions, especially in low-resource regimes. More qualitative analyses of hyper-parameters, training data size and alternative source datasets are shown in Appendix. 

\begin{table}[t]
\centering
\resizebox{0.49\textwidth}{!}{
\begin{tabular}{l||cc||cc}
\hline
\multirow{2}{*}{Model}               & \multicolumn{2}{c|}{Weibo-COVID19} & \multicolumn{2}{c}{Twitter-COVID19} \\ \cline{2-5} 
                                     & Acc.            & Mac-$\emph{F}_1$           & Acc.             & Mac-$\emph{F}_1$           \\ \hline \hline
$\text{BiGCN} (T)$          & 0.569           & 0.508            & 0.616            & 0.415            \\
$\text{BiGCN} (S)$          & 0.578           & 0.463            & 0.611            & 0.425            \\
$\text{BiGCN} (S,T)$ & 0.693           & 0.472            & 0.617            & 0.471            \\ \hline
DANN-BiGCN                           & 0.629           & 0.561            & 0.618            & 0.510            \\ \hline
CLR-BiGCN                       & 0.844           & 0.804            & 0.719            & 0.618            \\
ACLR-BiGCN                      & 0.873           & 0.861            & 0.765            & 0.686            \\ \hline
\end{tabular}}
\caption{Ablation studies on our proposed model.}
\vspace{-0.3cm}
\label{ablation}
\end{table}

\subsection{Early Detection}
Early alerts of rumors is essential to minimize its social harm. 
By setting detection checkpoints of ``delays" that can be either the count of reply posts or the time elapsed since the first posting, only contents posted no later than the checkpoints is available for model evaluation. The performance is evaluated by Macro F1 
obtained at each checkpoint. To satisfy each checkpoint, we incrementally scan test data in order of time until the target time delay or post volume is reached.

{\setlength{\abovecaptionskip}{-0.1cm}
\setlength{\belowcaptionskip}{-0.1cm}
\begin{figure}[t]
\centering
\subfigure{
\begin{minipage}[t]{0.5\linewidth}
\centering
\scalebox{0.75}{\includegraphics[width=5cm]{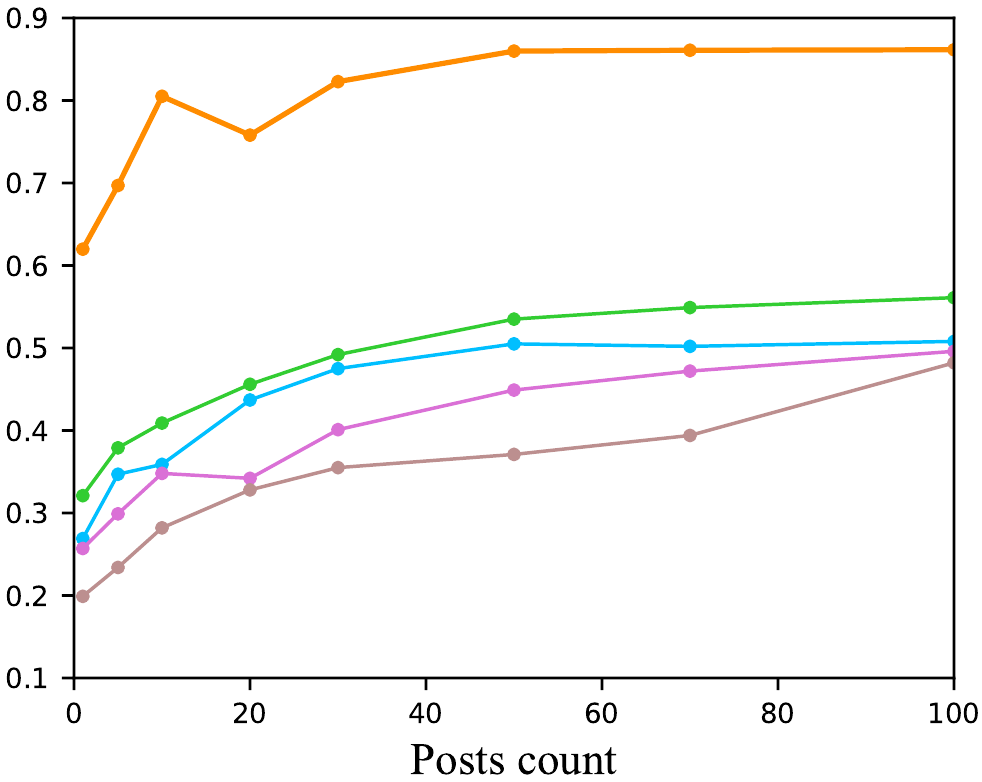}}
\label{fig:twitter15_early}
\end{minipage}%
}%
\subfigure{
\begin{minipage}[t]{0.5\linewidth}
\centering
\scalebox{0.75}{\includegraphics[width=5cm]{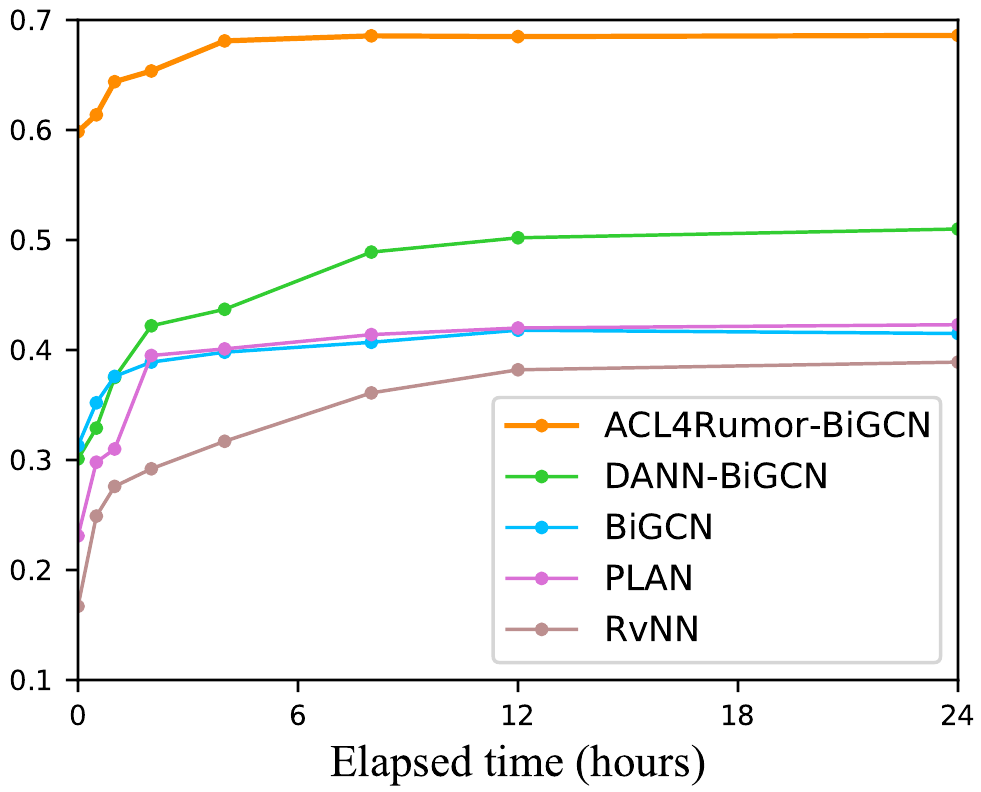}}
\label{fig:twitter16_early}
\end{minipage}%
}%
\centering
\caption{Early detection performance at different checkpoints of posts count (or elapsed time) on Weibo-COVID19 (left) and Twitter-COVID19 (right) datasets.}
\label{fig:early_detection}
\vspace{-0.2cm}
\end{figure}}

Figure~\ref{fig:early_detection} shows the performances of our approach versus DANN-BiGCN, BiGCN, PLAN, and RvNN at various deadlines. Firstly, we observe that our proposed ACLR-based approach outperforms other counterparts and baselines throughout the whole lifecycle, and reaches a relatively high Macro F1 score at a very early period after the initial broadcast. One interesting phenomenon is that the early performance of some methods may fluctuate more or less. It is because with the propagation of the claim there is more semantic and structural information but the noisy information is increased simultaneously. Our method only needs about 50 posts on Weibo-COVID19 and around 4 hours on Twitter-COVID19, to achieve the saturated performance, indicating the remarkably superior early detection performance of our method.

\subsection{Feature Visualization}

{\setlength{\abovecaptionskip}{-0.1cm}
\setlength{\belowcaptionskip}{-0.1cm}
\begin{figure}[t]
\centering
\subfigure{
\begin{minipage}[t]{0.5\linewidth}
\centering
\scalebox{0.75}{\includegraphics[width=5cm]{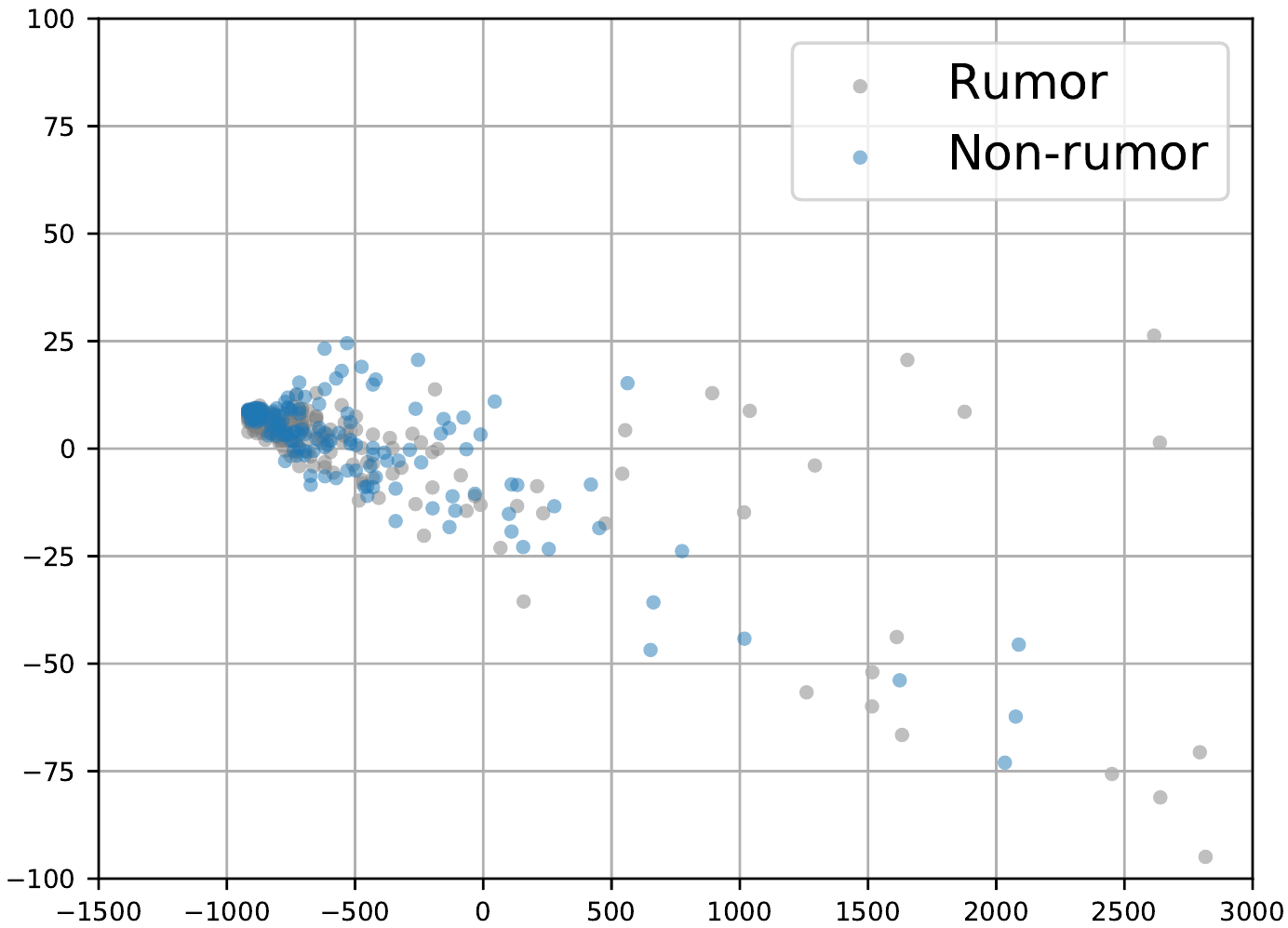}}
\label{fig:baseline_distribution}
\end{minipage}%
}%
\subfigure{
\begin{minipage}[t]{0.5\linewidth}
\centering
\scalebox{0.75}{\includegraphics[width=5cm]{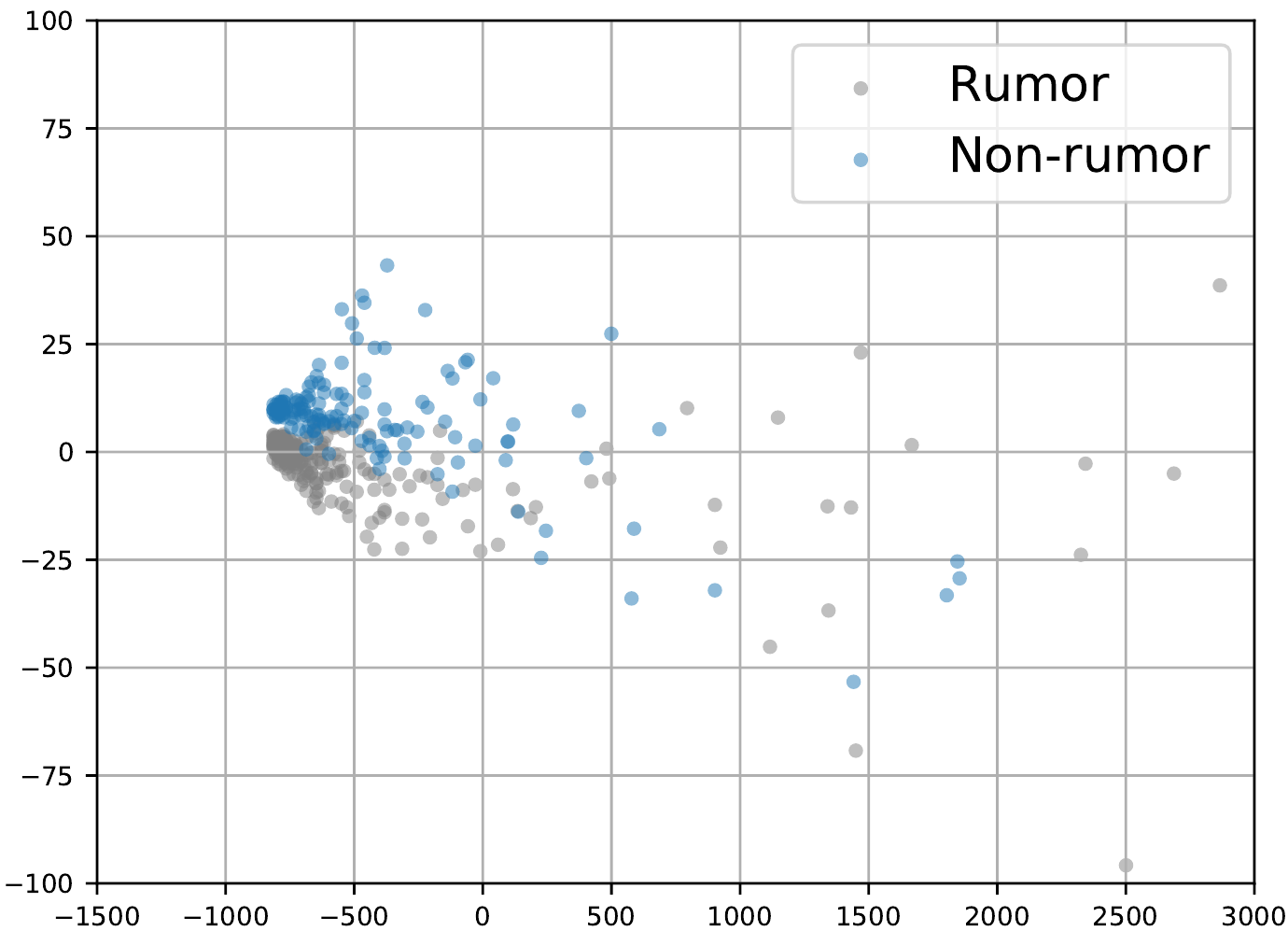}}
\label{fig:ACL_distribution}
\end{minipage}%
}%
\centering
\caption{Visualization of target event-level representation distribution.}
\label{fig:vis}
\vspace{-0.2cm}
\end{figure}}

Figure~\ref{fig:vis} shows the PCA visualization of learned target event-level features on BiGCN (left) and ACLR-BiGCN (right) for analysis. The left figure represents training with only classification loss, and the right figure uses ACLR for training. We observe that (1) due to the lack of sufficient training data, the features extracted with the traditional training paradigm are entangled, making it difficult to detect rumors in low-resource regimes; and (2) our ACLR-based approach learns more discriminative representations to improve low-resource rumor classification, reaffirming that our training paradigm can effectively transfer knowledge to bridge the gap between source and target data distribution resulting from different domains and languages.

\section{Conclusion and Future Work}
In this paper, we proposed a novel Adversarial Contrastive Learning framework to bridge low-resource gaps for rumor detection by adapting features learned from well-resourced data to that of the low-resource breaking events. Results on two real-world benchmarks confirm the advantages of our model in low-resource rumor detection task. In our future work, we plan to collect and apply our model on other domains and minority languages.

\section*{Acknowledgements}
This work was partially supported by HKBU One-off Tier 2 Start-up Grant (Ref. RCOFSGT2/20-21/SCI/004), HKBU direct grant (Ref. AIS 21-22/02) and MoE-CMCC ``Artificial Intelligence" Project No. MCM20190701.

\bibliography{anthology,custom}
\bibliographystyle{acl_natbib}

\clearpage
\appendix

\section{Datasets}
The focus of this work, as well as in many previous studies~\cite{ma2017detect, ma2018rumor, khoo2020interpretable, bian2020rumor}, is rumors on social media, not just the "fake news" strictly defined as a news article published by a news outlet that is verifiably false~\cite{shu2017fake,zubiaga2018detection}.
To our knowledge, there is no public dataset available for classifying propagation trees in tweets about COVID-19, where we need the tree roots together with the corresponding propagation structure, to be appropriately annotated with ground truth. In this paper, we organize and construct two datasets Weibo-COVID19 and Twitter-COVID19 for experiments. For Twitter-COVID19, the original dataset~\cite{kar2020no} of tweets was released with just the source tweet without its propagation thread. So we collected all the propagation threads using the Twitter academic API with the twarc2 package\footnote{\url{https://twarc-project.readthedocs.io/en/latest/twarc2_en_us/}} in python. Finally, we annotated the source tweets by referring to the labels of the events they are from the raw COVID-19 rumor dataset~\cite{kar2020no}, where rumors contain fact or misinformation to be verified while non-rumors do not. For Weibo-COVID19, data annotation similar to \citet{ma2016detecting}, a set of rumorous claims is gathered from the Sina community management center\footnote{\url{https://service.account.weibo.com/}} and non-rumorous claims by randomly filtering out the posts that are not reported as rumors. Weibo API is utilized to collect all the repost/reply messages towards each claim. Both Weibo-COVID19 and Twitter-COVID19 contain two binary labels: Rumor and Non-rumor. For Weibo-COVID19 as the target dataset, we use the TWITTER dataset~\cite{ma2017detect} as the source data in our low-resource (i.e., cross-domain and cross-lingual) settings; In terms of Twitter-COVID19 as the target dataset, we use WEIBO~\cite{ma2016detecting} as the source data. The statistics of the four datasets are shown in Table \ref{tab:statistics}. 

\begin{table*}[t]
\centering
\resizebox{0.9\textwidth}{!}{
\begin{tabular}{l||cc|cc}
\hline
Cross-Domain\&Lingual Settings & Source    & Target        & Source     & Target          \\ 
Statistics                      & TWITTER   & Weibo-COVID19 & WEIBO      & Twitter-COVID19 \\ \hline \hline
\# of events                    & 1154      & 399           & 4649       & 400             \\ \hline
\# of tree nodes                & 60409     & 26687         & 1956449    & 406185          \\ \hline
\# of non-rumors                & 579       & 146           & 2336       & 148             \\ \hline
\# of rumors                    & 575       & 253           & 2313       & 252             \\ \hline
Avg. time length/tree           & 389 Hours & 248 Hours     & 1007 Hours & 2497 Hours      \\ \hline
Avg. depth/tree                 & 11.67     & 4.31          & 49.85      & 143.03          \\ \hline
Avg. \# of posts/tree           & 52        & 67            & 420        & 1015            \\ \hline
Domain                          & Open      & COVID-19       & Open       & COVID-19         \\ \hline
Language                        & English   & Chinese       & Chinese    & English         \\ \hline
\end{tabular}}
\caption{Statistics of Datasets in Cross-Domain and Cross-Lingual Settings.}
\label{tab:statistics}
\end{table*}

\section{Implementation Details}
We set the number $L$ of the graph convolutional layer as 2, the trade-off parameter $\alpha$ as 0.5, and the adversarial perturbation norm $\epsilon$ as 1.5. The temperature $\tau$ is set to 0.1. Parameters are updated through back-propagation~\cite{collobert2011natural} with the Adam optimizer~\cite{loshchilov2018decoupled}. The learning rate is initialized as 0.0001, and the dropout rate is 0.2. Early stopping \cite{yao2007early} is applied to avoid overfitting. We run all of our experiments on one single NVIDIA Tesla T4 GPU. We set the total batch size to 64, where the batch size of source samples is set to 32, the same as target samples. The hidden and output dimensions of each node in the structure-based network are set to 512 and 128, respectively. Since the focus in this paper is primarily on better leveraging the contrastive learning for domain and language adaptation on top of event-level representations, we choose the $\text{XLM-R}_{\textit{Base}}$ (Layer number = 12, Hidden dimension = 768, Attention head = 12, 270M params) as our sentence encoder for language-agnostic representations at the post level. We use accuracy and macro-averaged F1 score, as well as class-specific F1 score as the evaluation metrics. Unusually, to conduct five-fold cross-validation on the target dataset in our low-resource settings, we use each fold (about 80 claim posts with propagation threads in the target data) in turn for training, and test on the rest of the dataset. The average runtime for our approach on five-fold cross-validation in one iteration is about 3 hours. The number of total trainable parameters is 1,117,954 for our model. We implement our model with pytorch\footnote{\url{pytorch.org}}.

\section{Qualitative Analysis}
\subsection{Effect of Adversarial Perturbation Norm}

Figure~\ref{fig:APN} shows the effect of adversarial perturbation norm on rumor detection performance. The X-axis denotes the value of $\epsilon$, where $\epsilon =0.0$ in the line means no adversarial augmentation. In general, the adversarial augmentation contributes to the improvements and $\epsilon \in [1.0, 2.0)$ achieves better performances. For the Weibo-COVID19 dataset, our proposed approach ACLR with a smaller adversarial perturbation can still obtain better results but lower than the results with an optimal range of perturbation, while large norms tend to damage the effect of ACLR. In terms of Twitter-COVID19, our method still performs well with a broad range of adversarial perturbations and the performance tends to stabilize as the norm value increases.

{
\begin{figure}[t]
    \centering
    \setlength{\belowcaptionskip}{0.1cm}
    \resizebox{0.45\textwidth}{!}{\includegraphics{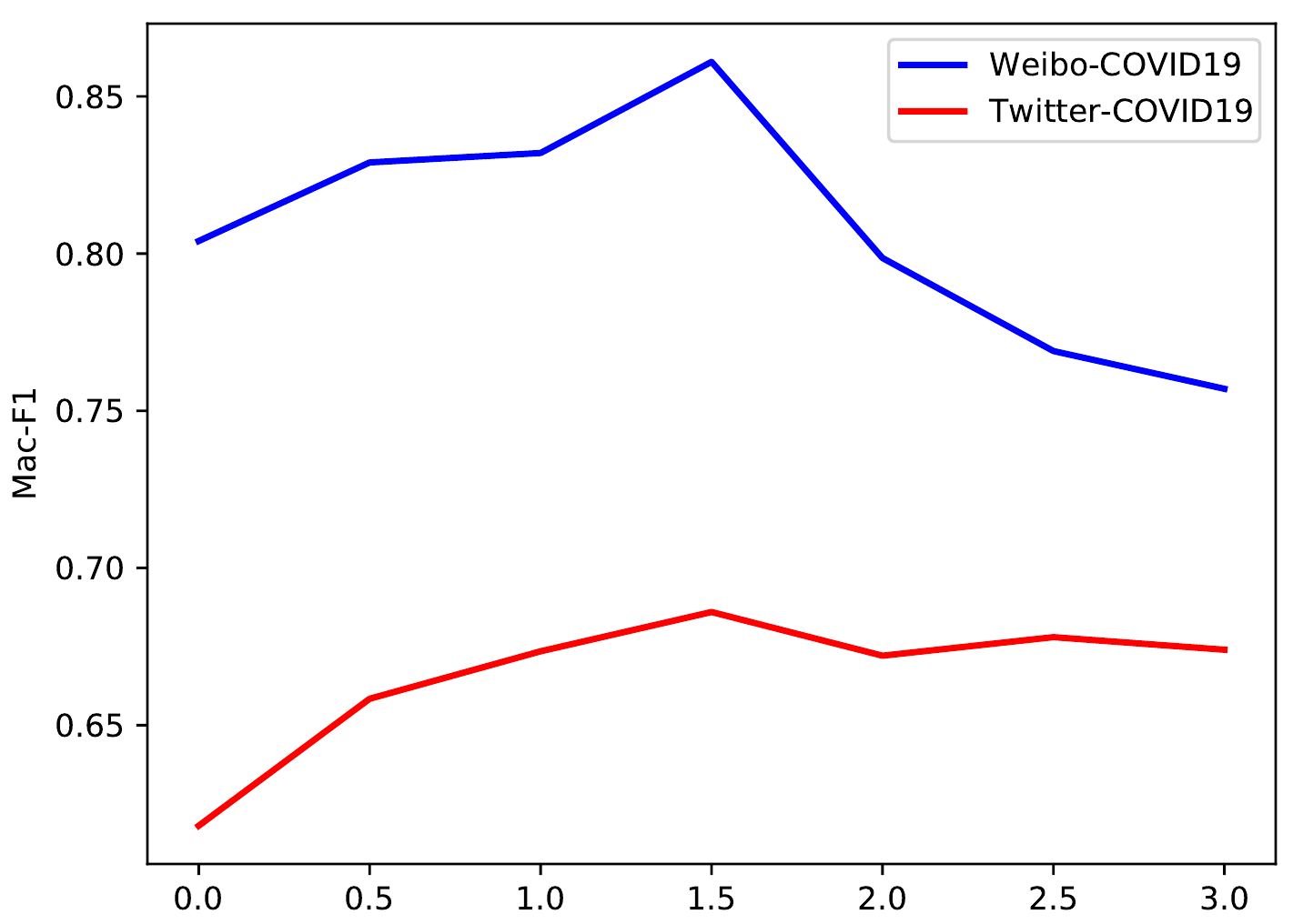}}
    \caption{Effect of Adversarial Perturbation Norm $\epsilon$.}
    \label{fig:APN}
\end{figure}}

{
\begin{figure}[t]
    \centering
    \setlength{\belowcaptionskip}{0.1cm}
    \resizebox{0.45\textwidth}{!}{\includegraphics{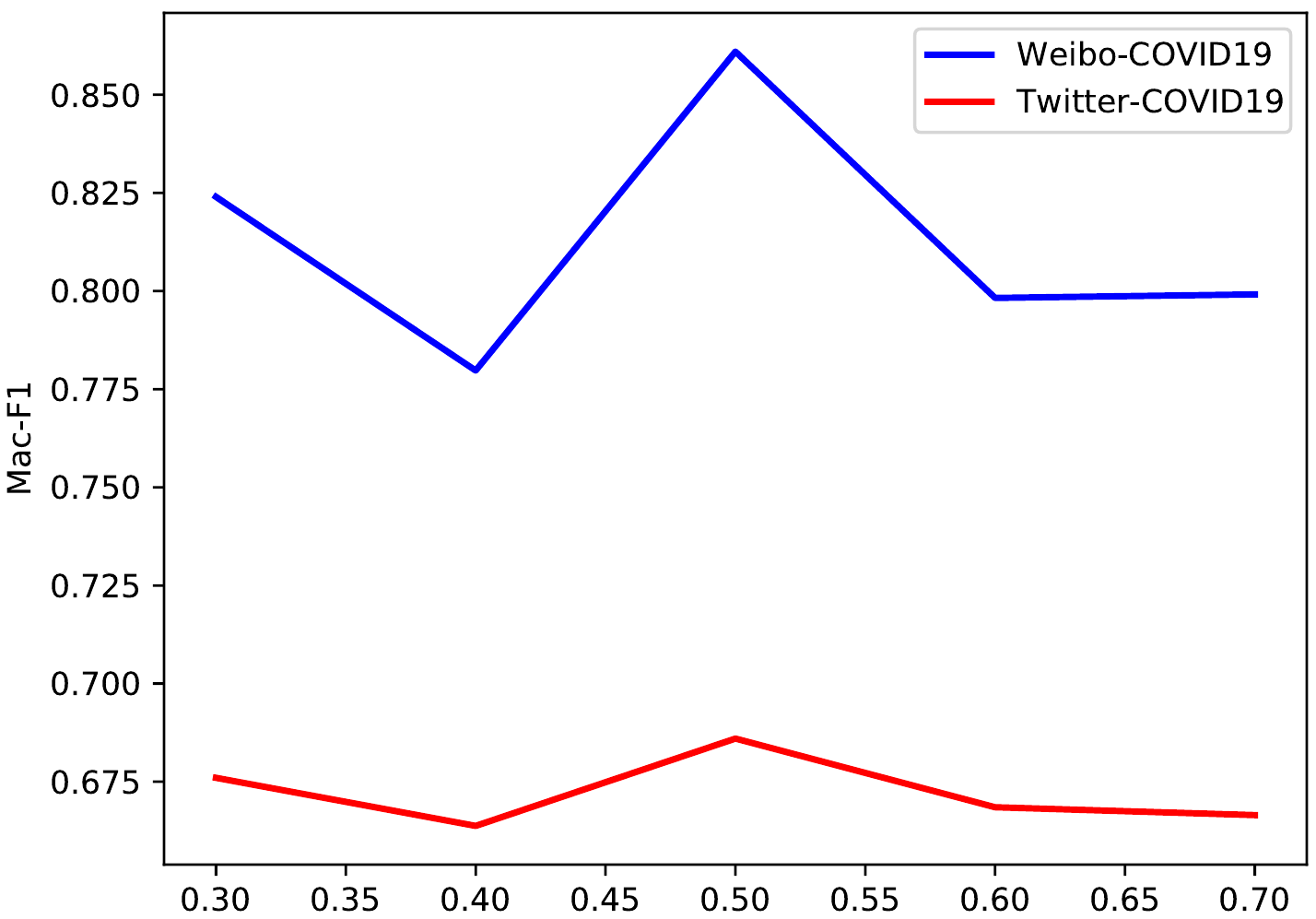}}
    \caption{Effect of trade-off parameter $\alpha$.}
    \label{fig:alpha}
\end{figure}}

\subsection{Effect of Trade-off Parameter between Classification and Contrastive Objectives}
To study the effects of the trade-off hyper-parameter in our training paradigm, we conduct ablation analysis under ACLR architecture (Figure~\ref{fig:alpha}). We can see that $\alpha = 0.5$ achieves the best performance while the point where $\alpha =0.3$ also has good performance. Looking at the overall trend, the performance fluctuates more or less as the value of $\alpha$ grows. We conjecture that this is because the supervised contrastive objective, while optimizing the representation distribution, compromises the mapping relationship with labels. Multitask means optimizing two losses simultaneously. This setting leads to mutual interference between two tasks, which affects the convergence effect. This phenomenon points out the direction for our further research in the future.

\subsection{Effect of Target Training Data Size.}
Figure~\ref{fig:training_size} shows the effect of target training data size. We randomly choose training data with a certain proportion from target data and use the rest set for evaluation. We use the cross-domain and cross-lingual settings concurrently for model training, the same as the main experiments. Results show that with the decrease of training data size, the performance gradually decreases. Especially for Weibo-COVID19, it will be greatly affected. However, even when only 20 target data are used for training, our model can still achieve more than approximately 60\% and 65\% rumor detection performance (Macro F1 score) on two target data sets Weibo-COVID19 and Twitter-COVID19 respectively, which further proves ACLR has strong applicability for improving low-resource rumor detection on social media.

{
\begin{figure}[t]
    \centering
    \setlength{\belowcaptionskip}{0.1cm}
    \resizebox{0.45\textwidth}{!}{\includegraphics{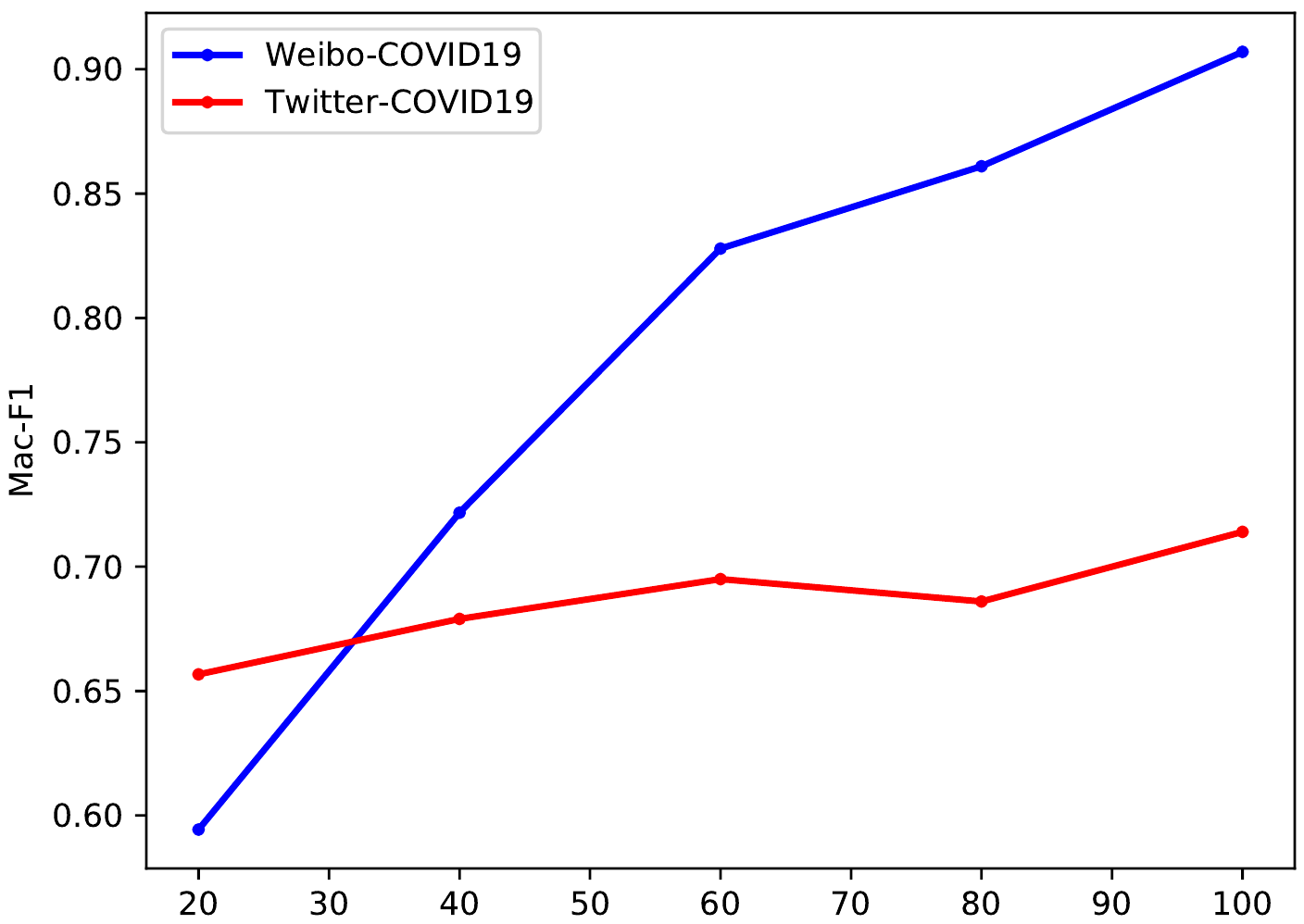}}
    \caption{Effect of target training data size.}
    \label{fig:training_size}
\end{figure}}

\subsection{Discussion about Low-Resource Settings}

In this section, we evaluate our proposed framework with different source datasets to discuss the low-resource settings in our experiments. Considering the cross-domain and cross-lingual settings in the main experiments, we also conduct an experiment in cross-domain settings. Specifically, for the Weibo-COVID as the target data, we utilize the WEIBO dataset as the source data with rich annotation. In terms of Twitter-COVID19, we set the TWITTER dataset as the source data. Table~\ref{tab:low-resource} depicted the results in different low-resource settings. It can be seen from the results that our model performs generally better in cross-domain and cross-lingual settings concurrently than that only in cross-domain settings, which demonstrates the key insight to bridge the low-resource gap is to relieve the limitation imposed by the specific language resource dependency besides the specific domain. Our proposed adversarial contrastive learning framework could alleviate the low-resource issue of rumor detection as well as reduce the heavy reliance on datasets annotated with specific domain and language knowledge.

\begin{table}[t]
\centering
\resizebox{0.46\textwidth}{!}{
\begin{tabular}{l|cc|cc}
\hline
Target     & \multicolumn{2}{c|}{Weibo-COVID19} & \multicolumn{2}{c}{Twitter-COVID19} \\ \hline
Settings   & Acc.             & Mac-$\emph{F}_1$          & Acc.             & Mac-$\emph{F}_1$           \\ \hline
Cross-D\&L & 0.873            & \textbf{0.861}  & \textbf{0.765}   & \textbf{0.686}   \\ \hline
Cross-D    & \textbf{0.884}   & 0.855             & 0.737            & 0.623            \\ \hline
\end{tabular}}
\caption{Rumor detection results of our proposed framework in different low-resource settings. Cross-D\&L denotes the cross-domain and cross-lingual settings and Cross-D denotes the cross-domain and monolingual settings.}
\label{tab:low-resource}
\end{table}

\section{Future Work}
We will explore the following directions in the future:
\begin{enumerate}
\item We are going to explore the pre-training method with contrastive learning and then finetune the model with classification loss, which may further improve the performance and stability of the model.
\item Considering that our model has explicitly overcome the restriction of both domain and language usage in different datasets, we plan to evaluate our model on the datasets about more breaking events in low-resource domains and/or languages by leveraging existing datasets with rich annotation. We believe that our work could provide new guidance for future rumor detection about breaking events on social media.

\end{enumerate}

\end{document}